\documentclass{article}

\usepackage[utf8]{inputenc} 
\usepackage[T1]{fontenc}    
\usepackage{booktabs}       
\usepackage{amsfonts, amsmath, amssymb}       
\usepackage{nicefrac}       
\usepackage{microtype}      
\usepackage{hyperref}

\usepackage[toc,page,header]{appendix}
\usepackage{minitoc}
\usepackage{comment}

\PassOptionsToPackage{numbers, compress}{natbib}

\usepackage[final]{neurips_2020}

\usepackage{graphicx}

\usepackage{subcaption}

\usepackage{bm}

\usepackage[ruled]{algorithm2e}

\usepackage{multicol}
\usepackage{mathtools}

\newcommand{\E}{\mathbb{E}}

\usepackage{color}

\usepackage{xcolor}
\definecolor{dark-red}{rgb}{0.4,0.15,0.15}
\definecolor{dark-blue}{rgb}{0.15,0.15,0.4}
\definecolor{medium-blue}{rgb}{0,0,0.5}
\hypersetup{
   colorlinks, linkcolor={dark-blue},
   citecolor={dark-blue}, urlcolor={medium-blue}
}

\title{Learning Invariances in Neural Networks}

\author{Gregory Benton \quad Marc Finzi \quad Pavel Izmailov \quad Andrew Gordon Wilson\\
Courant Institute of Mathematical Sciences\\
New York University}

\begin{document}

\maketitle

\begin{abstract}
Invariances to translations have imbued convolutional neural networks with powerful generalization properties. However, we often do not know a priori what invariances are present in the data, or to what extent a model should be invariant to a given symmetry group. We show how to \emph{learn} invariances and equivariances by parameterizing a distribution over augmentations and optimizing the training loss simultaneously with respect to the network parameters and augmentation parameters. With this simple procedure we can recover the correct set and extent of invariances on image classification, regression, segmentation, and molecular property prediction from a large space of augmentations, on training data alone. 
\end{abstract}

\section{Introduction}

The ability to learn constraints or symmetries is a foundational property of intelligent systems. Humans are able to discover patterns and regularities in data that provide compressed representations of reality, such as translation, rotation, intensity, or scale symmetries. 
Indeed, we see the value of such constraints in deep learning. Fully connected networks are more flexible than convolutional networks, but convolutional networks are more broadly impactful because they enforce the \emph{translation equivariance} symmetry: when we translate an image, the outputs of a convolutional layer translate in the same way \citep{lecun1998convolutional, cohen2016group}. Further gains have been achieved by recent work hard-coding additional symmetries, such as rotation equivariance, into convolutional neural networks \citep[e.g.,][]{cohen2016group,worrall2017harmonic,zhou2017oriented,marcos2017rotation}

But we might wonder whether it is possible to \emph{learn} that we want to use a convolutional neural network.
Moreover, we typically do not know which constraints are suitable for a given problem, and to what extent those constraints should be enforced. The class label for the digit `6' is rotationally invariant up until it becomes a `9'. Like biological systems, we would like to automatically discover the appropriate symmetries. This task appears daunting, because standard learning objectives such as maximum likelihood select for flexibility, rather than constraints \citep{mackay2003information, minka2001automatic}. 

In this paper, we provide an extremely simple and practical approach to automatically discovering invariances and equivariances, \emph{from training data alone}. Our approach operates by learning a distribution over augmentations, then training with augmented data, leading to the name \emph{Augerino}. 
Augerino (1) can learn both invariances and equivariances over a wide range of symmetry groups, including translations, rotations, scalings, and shears; (2) can discover partial symmetries, such as rotations not spanning the full range from $[-\pi, \pi]$;  (3) can be combined with any standard architectures, loss functions, or optimization algorithm with little overhead; (4) performs well on regression, classification, and segmentation tasks, for both image and molecular data. 

To our knowledge, Augerino is the first approach that can learn symmetries in neural networks from training data alone, without requiring a validation set or a special loss function. In Sections \ref{sec: augerino}-\ref{sec: why} we introduce Augerino and show why it works. The accompanying code can be found at \url{https://github.com/g-benton/learning-invariances}.

\section{Related Work}

There is a large body of work constructing convolutional neural networks that have \emph{hard-coded} invariance or equivariance to a set of transformations, such as rotation \citep{cohen2016group,worrall2017harmonic,zhou2017oriented,marcos2017rotation} and scaling \citep{worrall2019deep,sosnovik2019scale}. While recent methods use a representation theoretic approach to find a basis of equivariant convolutional kernels \citep{cohen2016steerable,worrall2017harmonic,weiler2019general}, the older method of 
\citet{laptev2016ti} pools network outputs over many hard-coded transformations of the input for fixed invariances, but does not consider equivariances or learning the transformations.

\citet{van2018learning} learn transformations for learning invariances in kernel methods from training data, using the marginal likelihood of a Gaussian process. The marginal likelihood, which is the  integral of the product of the likelihood with a parameter prior, automatically selects for constraints \citep[e.g.,][]{mackay2003information}. They propose a similar pipeline of learning the parameters of a transformation directly by backpropagation and the reparametrization trick. In contrast to their work, we develop a framework that can be easily applied to deep neural networks with standard loss functions, without needing to compute a marginal likelihood (which is typically intractable). Our framework can also learn more general transformations through the exponential map, as well as \emph{equivariant} models. 

With a desire to automate the machine learning pipeline, \citet{cubuk2019autoaugment} introduced \emph{AutoAugment} in which reinforcement learning is used to find an optimal augmentation policy within a discrete search space. At the expense of a massive computational budget for the search, AutoAugment brought substantial gains in image classification performance, including state-of-the-art results on ImageNet. The AutoAugment framework was extended first to \emph{Fast AutoAugment} in \citet{lim2019fast}, improving both the speed and accuracy of AutoAugment by using Bayesian data augmentation \citep{tran2017bayesian}. Both \citet{cubuk2019autoaugment} and \citet{lim2019fast} apply a reinforcement learning approach to searching the space of augmentations, significantly differing from our work which directly optimizes distributions over augmentations with respect to the training loss. 

\emph{Faster AutoAugment} \citep{hataya2019faster}, which uses a GAN framework to match augmentations to the data distribution, and \emph{Differentiable Automatic Data Augmentation} \citep{li2020dada} which applies a DARTS \citep{liu2018darts} bi-level optimization procedure to learn augmentation from the validation loss are most similar to Augerino in the discovery of distributions over augmentations. Both methods learn augmentations from data using the reparametrization trick; however unlike \citet{li2020dada} and \citet{liu2018darts}, we learn augmentations directly from the training loss without need for GAN training or the complex DARTS procedure \citep{liu2018darts,xu2019pc,liang2019darts}, and are specifically learning degrees of invariances and equivariances. 

To the best of our knowledge, Augerino is the first work to \emph{learn} invariances and equivariances in neural networks from training data alone. The ability to automatically discover symmetries enables us to uncover interpretable salient structure in data, and provide better generalization.

\section{Augerino: Learning Invariances through Augmentation}
\label{sec: augerino}

A simple way of constructing a model invariant to a given group of transformations is to average the outputs of an arbitrary model for the inputs transformed with all the transformations in the group.
For example, if we wish to make a given image classifier invariant to horizontal reflections, we can average the predictions of the network for the original and reflected input.

\begin{figure}[t]
    \centering
    \includegraphics[width=0.8\linewidth]{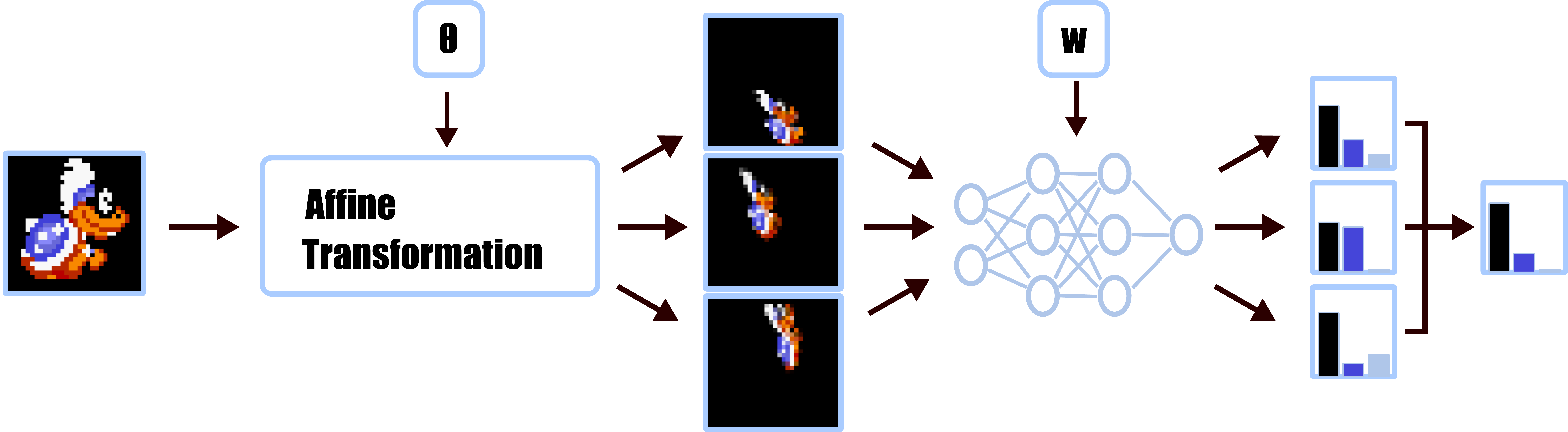}
    \caption{The Augerino framework. Augmentations are sampled from a distribution governed by parameters $\theta$, and applied to an input to produce multiple augmented inputs. These augmented inputs are then passed to a neural network with weights $w$, and the final prediction is generated by averaging over the multiple outputs. Augerino discovers invariances by learning $\theta$ from training data alone.}
    \label{fig: augerino-diagram}
\end{figure}

Augerino functions by sampling multiple augmentations from a parametrized distribution then applying these augmentations to an input to acquire multiple augmented samples of the input. The augmented input samples are each then passed through the model, with the final prediction being generated by averaging over the individual outputs. We present the Augerino framework in Figure \ref{fig: augerino-diagram}.

Now, suppose we are working with a set $\mathcal S$ of transformations. Relevant transformations may not always form a group structure, such as rotations $R_\phi$ by limited angles in the range $\phi \in [-\theta, \theta]$.
Given a neural network $f_w$, with parameters $w$,  
we can make a new model $\bar{f}$ which is approximately invariant to transformations $\mathcal S$ by averaging the outputs over a uniform distribution $\mu_\theta(\cdot)$ over the transformations $g \in S$ with $\textrm{supp}(\mu_\theta) = \mathcal{S}$\footnote{See Appendix \ref{app: invariance} for further discussion on forming the invariant model.}
\citep[e.g.,][]{laptev2016ti,raj2017local,van2018learning}
:
\begin{equation}\label{eqn: aug-expectation}
    \bar{f}(x) = \E_{g \sim \mu_\theta} f(gx).
\end{equation}

Since the cross-entropy loss $\ell$ for classification is linear in the class log probabilities, we can pull the expectation outside of the loss:
\begin{equation}\label{eq:augerino_expected_loss}
    \ell(\bar{f}(x)) = \ell(\E_{g \sim \mu_\theta} f(gx)) = \E_{g \sim \mu_\theta} \ell(f(gx)).
\end{equation}
As stochastic gradient descent only requires an unbiased estimator of the gradients, we can train the augmentation averaged model $\bar{f}$ \emph{exactly} by minimizing the loss of $f(gx)$ averaged over a finite number of samples from $g\sim \mu_\theta$ at training time, using a Monte Carlo estimator.

To learn the invariances we can also backpropagate through to the parameters $\theta$ of the distribution $\mu_\theta$ by using the reparametrization trick \citep{kingma2013auto}. For example, for a uniform distribution over rotations with angles $U[-\theta,\theta]$, we can parametrize the rotation angle by $\phi = \theta\epsilon$ with $\epsilon \sim U[-1,1]$.
The loss $L(\cdot)$ for the augmentation-averaged model on an input $x$ can be computed as
\begin{equation}\label{eq:augerino_loss}
    L_x(\theta, w) = \E_{\phi \sim U[-\theta, \theta]} \ell\big(f_w(R_\phi x)\big) = \E_{\epsilon \sim U[-1, 1]}\ell\big( f_w(R_{\epsilon \theta}x)\big).
\end{equation}

Specifically, during training we can use a single sample from the augmentation distribution to estimate the gradients.
The learned range of rotations $[-\theta, \theta]$ would correspond to the extent rotational invariance is present in the data. 
With a more general set of $k$ transformations, we can similarly define a distribution $\mu_\theta(\cdot)$ over the transformation elements using the reparametrization trick $g = g_\epsilon = \epsilon \odot \theta$, with $\epsilon \sim U[-1,1]^k$ and $\theta \in \mathbb{R}^k$. The reparametrized loss is then
\begin{equation}\label{eq:augerino_loss}
    L_x(\theta, w) = \E_{\epsilon \sim U[-1, 1]^k} \ell\big(f_w(g_\epsilon x)\big).
\end{equation}
In Section \ref{sec:affine} we describe a parameterization of the set of affine transformations which includes translations, rotations, and scalings of the input as special cases.
In this fashion, we can train both the parameters of the augmentation averaged model $\bar{f}$ consisting both of the weights $w$ of $f_w$ and the parameters $\theta$ of the augmentation distribution $\mu_\theta$.

\paragraph{Test-time Augmentation} 
At test time we sample multiple transformations $g \sim \mu_\theta$ and make a prediction by averaging over the predictions generated for each transformed input, approximating the expectation in Equation \eqref{eqn: aug-expectation}. We 
further discuss train and test time augmentation in Appendix \ref{app: training-details}.

\paragraph{Regularized Loss}
Invariances correspond to constraints on the model, and in general the most unconstrained model may be able to achieve the lowest training loss. 
However, we have a prior belief that a model should preserve \emph{some} level of invariance, even if standard losses cannot account for this preference. To bias training towards solutions that incorporate invariances, we add a regularization penalty to the network loss function that promotes broader distributions over augmentations.
Our final loss function is given by
\begin{equation}
    \label{eq:augerino_final_loss}
    L_x(\theta, w) = \E_{g \sim \mu_\theta}  \ell\big(f_w(g x)\big) + \lambda R(\theta),
\end{equation}
where $R$ is a regularization function encouraging coverage of a larger volume of transformations and $\lambda$ is the regularization weight (the form of $R(\theta)$ is discussed in Section \ref{sec:affine}). 
In practice we find that the choice of $\lambda$ is \emph{largely unimportant}; the insensitivity to the choice of $\lambda$ is demonstrated throughout Sections \ref{sec: shades} and \ref{sec: boosting} in which performance is consistent for various values of $\lambda$. This is due to the fact that there is essentially no gradient signal for $\theta$ over the range of augmentations consistent with the data, so even a small push is sufficient. We discuss further why Augerino is able to learn the correct level of invariance --- \emph{without sensitivity to $\lambda$, and from training data alone} --- in Section~\ref{sec: why}.

We refer to the introduced method as \textit{Augerino}. We summarize the method in Algorithm \ref{alg:augerino}. 

\begin{algorithm}[H]
\SetAlgoLined
\textbf{Inputs:} \\
Dataset $\mathcal D$; 
parametric family $g$ of data augmentations and a distribution $\mu_\theta$ over the parameters $\theta$; 
neural network $f_w$ with parameters $w$;
number $n_\text{copies}$ of augmented inputs to use during training;
number of training steps $N$.
\\
\For{$i = 1, \dots, N$}{
    Sample a mini-batch $\tilde x$ from $\mathcal D$; \\
    For each datapoint in $\tilde x$ sample $n_\text{copies}$ transformations from $\mu_\theta$; \\
    Average predictions of the network $f_w$ over $n_\text{copies}$ data transformations of $\tilde x$;\\
    Compute the loss \eqref{eq:augerino_final_loss}, $L_{\tilde x}(\theta, w)$ using the averaged predictions;\\
    Take the gradient step to update the parameters $w$ and $\theta$; \\
}
\caption{Learning Invariances with Augerino}
\label{alg:augerino}
\end{algorithm}

\subsection{Extension to Equivariant Predictions}
\label{sec:equivariance}

We now generalize Augerino to problems where the targets are \textit{equivariant} rather than invariant to a certain set of transformations.
We say that target values are equivariant to a set of input transformations if the targets for a transformed input are transformed in the same way as the input. Formally, a function $f$ is equivariant to a symmetry transformation $g$, if applying $g$ to the input of the function is the same as applying $g$ to the output, such that $f(gx) = gf(x)$.
For example, in image segmentation if the input image is rotated the target segmentation mask should also be rotated by the same angle, rather than being unchanged.

To make the Augerino model equivariant to transformations sampled from $\mu_{\theta}(\cdot)$, we can average the inversely transformed outputs of the network for transformed inputs: 
\begin{equation}
    \label{eq:augerino_equivariant_loss}
    f_{\text{aug-eq}}(x) = \E_{g \sim \mu_\theta} g^{-1}f(g x).
\end{equation}
Supposing that $g$ acts linearly on the image then the model is equivariant:
\begin{align}
    f_{\text{aug-eq}}(h x) &= \E_{g \sim \mu_\theta} g^{-1} f(ghx) 
                                   = \E_{g \sim \mu_\theta}h (gh)^{-1}f(gh x) 
                                   = h \E_{u \sim \mu_\theta} u^{-1} f(u x) \\
     &=  h f_{\text{aug-eq}}(x)
\end{align}
where $u = gh$ and the distribution is right invariant: for any measurable set $S$, $\forall h\in G: \mu_{\theta}(S) = \mu_{\theta}(hS)$. If the distribution over the transformations is uniform then the model is equivariant. 

\subsection{Parameterizing Affine Transformations}
\label{sec:affine}

We now show how to parametrize a distribution over the set of affine transformations of $2d$ data (e.g. images). 
With this parameterization, Augerino can learn from a broad variety of augmentations including translations, rotations, scalings and shears.

The set of affine transformations form an algebraic structure known as a Lie Group. To apply the reparametrization trick, we can parametrize elements of this Lie Group in terms of its Lie Algebra via the exponential map \cite{falorsi2019reparameterizing}.
With a very simple approach, we can define bounds $\theta_i$ on a uniform distribution over the different exponential generators $G_i$ in the Lie Algebra:
\begin{equation}\label{eqn: aug-generator}
    g_\epsilon = \textrm{exp}\left(\sum_{i} \epsilon_i \theta_i G_i\right) \quad \epsilon \sim U[-1,1]^k, 
\end{equation}
where $\textrm{exp}$ is the matrix exponential function:
$\textrm{exp}(A) = \sum_{n=0}^\infty \frac 1 {n!} A^n$. \footnote{Mathematically speaking, this distribution is a \textit{pushforward} by the $\textrm{exp}$ map of a scaled cube with side lengths $\theta_i$ of a cube $\mu_\theta(\cdot) = \textrm{exp}_*\mathrm{Cube}_\theta(\cdot)$.}%

The generators of the affine transformations in $2d$, $G_1, \dots, G_6$, correspond to translation in $x$, translation in $y$, rotation, scaling in $x$, scaling in $y$, and shearing; we write out these generators in Appendix \ref{app: generating-matrices}. The exponential map of each generating matrix produces an affine matrix that can be used to transform the coordinate grid points of the input like in \citet{jaderberg2015spatial}. 
To ensure that the parameters $\theta_i$ are positive, we learn parameters $\tilde{\theta}_i$ where $\theta_i = \log(1 + \exp{\tilde{\theta}_i})$.
In maximizing the volume of transformations covered, it would be geometrically sensible to maximize the Haar measure $\mu_H(S)$ of the set of transformations $S = \mathrm{exp}(\mathrm{Cube}_\theta)$ that are covered by Augerino, which is similar to the volume covered in the Lie Algebra $\mathrm{Vol}(\mathrm{Cube}_\theta) = \Pi_{i=1}^k \theta_i$. However, we find that even the negative $L_2$ regularization $R(\theta) = -\|\theta\|^2$ on the bounds $\theta_i$
is sufficient to bias the model towards invariance.
More intuitively, the regularization penalty biases solutions towards values of $\theta$ which induce broad distributions over affine transformations, $\mu_\theta$.

 We apply the $L_2$ regularization penalty on both classification and regression problems, using cross entropy and mean squared error loss, respectively. This regularization method is effective, interpretable, and leads to the discovery of the correct level of invariance for a wide range of $\lambda$.

\section{Shades of Invariance}\label{sec: shades}

We can broadly classify invariances in three distinct ways: first there are cases in which we wish to be completely invariant to transformations in the data, such as to rotations on the rotMNIST dataset. 
There are also cases in which we want to be only partially invariant to transformations, i.e. \emph{soft} invariance, such as if we are asking if a picture is right side up or upside down. 
Lastly, there are cases in which we wish there to be no invariance to transformations, such as when we wish to predict the rotations themselves. 
We show that Augerino can learn full invariance, soft invariance, and no invariance to rotations. We then explain in Section~\ref{sec: why} why Augerino is able to discover the correct level of invariance from training data alone. Incidentally, soft invariances are the most representative of real-world problems, and also the most difficult to correctly encode a priori --- where we most need to learn invariances.

For the experiments in this and all following sections we use a $13$-layer CNN architecture from \citet{laine2016temporal}. 
We compare Augerino trained with three values of $\lambda$ from Equation \ref{eq:augerino_final_loss}; $\lambda = \{0.01, 0.05, 0.1\}$ corresponding to low, standard, and high levels of regularization. 
To further emphasize the need for invariance to be \emph{learned} as opposed to just embedded in a model we also show predictions generated from an invariant $E(2)$-steerable network \citep{cohen2016steerable}.
Specific experimental and training details are in Appendix \ref{app: training-details}.

\subsection{Full Rotational Invariance: rotMNIST}
\begin{figure*}%
    \centering
    \includegraphics[width=0.9\linewidth]{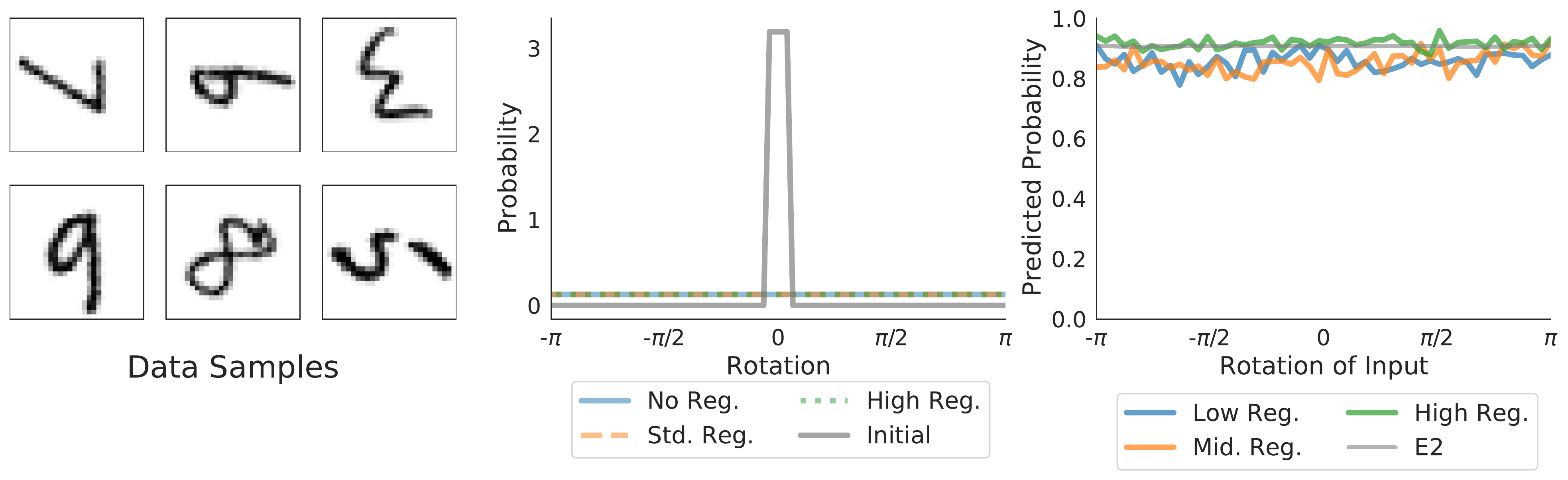}
    \caption{\textbf{Left:} Samples of the rotated digits in the data. 
    \textbf{Center:} The initial and learned distributions over rotations. 
    \textbf{Right:} The prediction probabilities of the correct class label over rotated versions of an image; the model learns to be approximately invariant to rotations under all levels of regularization.}
    \label{fig: rotmnist}
\end{figure*}

The rotated MNIST dataset (rotMNIST) consists of the MNIST dataset with the input images randomly rotated. As the dataset has an inherent augmentation present (random rotations), we desire a model that is invariant to such augmentations. With Augerino, we aim to approximate invariance to rotations by learning an augmentation distribution that is uniform over all rotations in $[0, 2\pi]$. 

Figure \ref{fig: rotmnist} shows the learned distribution over rotations to apply to images input into the model. On top of learning the correct augmentation through automatic differentiation using \emph{only} the training data, we achieve $98.9\%$ test accuracy. We also see the level of regularization has little effect on performance.
To our knowledge, only \citet{weiler2019general} achieve better performance on the rotMNIST dataset, using the correct equivariance already hard-coded into the network.

\subsection{Soft Invariance: Mario \& Iggy}
\label{sec:soft}

\begin{figure*}
    \centering
    \includegraphics[width=0.8\linewidth]{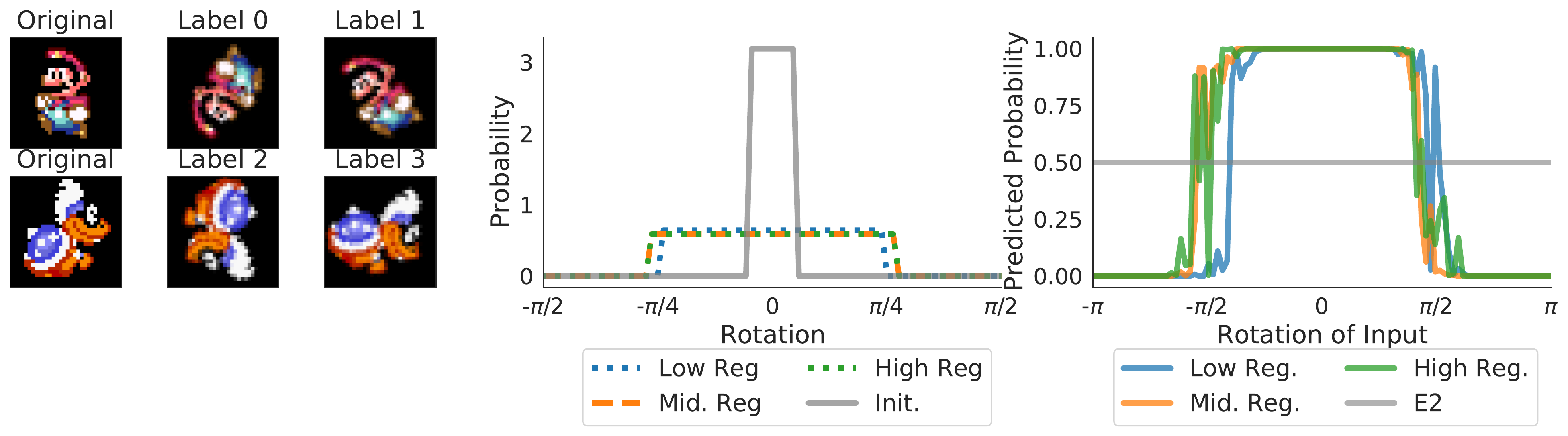}

    \caption{\textbf{Left:} Example data from the constructed Mario dataset. Labels are dependent on both the character, Mario or Iggy, and the rotation, upper half- or lower half-plane.
    \textbf{Center:} The initial and learned distribution over rotations. Rotations in the data are limited to $[-\pi/4, \pi/4]$ and $[-\pi, -3\pi/4] \cup [3\pi/4, \pi]$, meaning that augmenting an image by no more than $\pi/4$ radians will keep the rotation in the same half of the plane as where it started.
    The learned distributions approximate the invariance to rotations in $[-\pi/4, \pi/4]$ that is present in the data.
    \textbf{Right:} The predicted probability of label $1$ for input images of Mario rotated at various angles. $E2$-steerable model is invariant, and incapable of distinguishing between inputs of different rotations.}
    \label{fig: mario}
\end{figure*}

We show that Augerino can learn \emph{soft} invariances --- e.g. invariance to a subset of transformations such as only partial rotations. To this end, we consider a dataset in which the labels are dependent on both image and pose. 
We use the sprites for the characters Mario and Iggy from Super Mario World, randomly rotated in the intervals of $[-\pi/4, \pi/4]$ and $[-\pi, -3\pi/4] \cup [3\pi/4, \pi]$ \cite{smw}.  
There are $4$ labels in the dataset, one for the Mario sprite in the upper half plane, one for the Mario sprite in the lower half plane, one for the Iggy sprite in the upper half plane, and one for the Iggy sprite in the lower half plane; we show an example demonstrating each potential label in Figure \ref{fig: mario}.

In Figure \ref{fig: mario}, the limited rotations present in the data give that the labels are invariant to rotations of up to $\pi/4$ radians. Augerino learns the correct augmentation distribution, and the predicted labels follow the desired invariances
to rotations in $[-\pi/4, \pi/4]$.  

\begin{figure*}%
    \centering
    \includegraphics[width=0.9\linewidth]{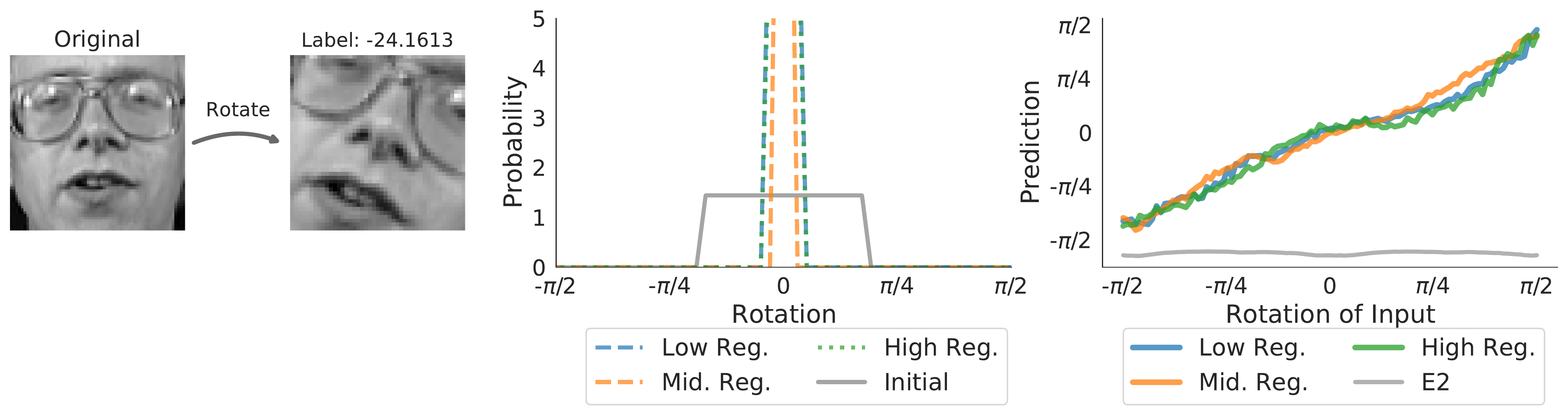}

    \caption{\textbf{Left:} The data generating process for the Olivetti faces dataset. The labels correspond to the rotation of the input image. 
    \textbf{Center:} The initialized and learned distributions over rotations.
    \textbf{Right:} The predictions generated as an input is rotated. Here we see that there is no invariance present for any level of regularization - as the image rotates the predicted label changes accordingly. The $E2$-steerable network fails for this task, as the invariance to rotations prevents us from being able to predict the rotation of the image.}
    \label{fig: olivetti}
\end{figure*}

\subsection{Avoiding Invariance: Olivetti Faces}

To test that Augerino can avoid unwanted invariances we train the model on the rotated Olivetti faces dataset \citep{hinton2008using}. 
This dataset consists of $10$ distinct images of $40$ different people. We select the images of $30$ people to generate the training set, randomly rotating each image in $[-\pi/2, \pi/2]$, retaining the angle of rotation as the new label. We then crop the result to $45 \times 45$ pixel square images. We repeat the process $30$ times for each image, 
generating $9000$ training images. Figure \ref{fig: olivetti} shows the data generating process and the corresponding label. Augmenting the image with any rotation would make it impossible to learn the angle by which the original image was rotated.

We find experimentally in Figure \ref{fig: olivetti} that when we initialize the Augerino model such that the distribution over the rotation generating matrix $G_3$ is uniform $[0, 1]$, training for $200$ epochs reduces the distribution on the rotational augmentation to have domain of support $0.003$ radians wide. The model learns a nearly fixed transformation in each of the $5$ other spaces of affine transformation, all with domains of support for the weights $w_i$ under $0.1$ units wide.

\section{Why Augerino Works}\label{sec: why}

\begin{figure*}
    \centering
    \begin{tabular}{c}
         \includegraphics[height=0.22\textwidth]{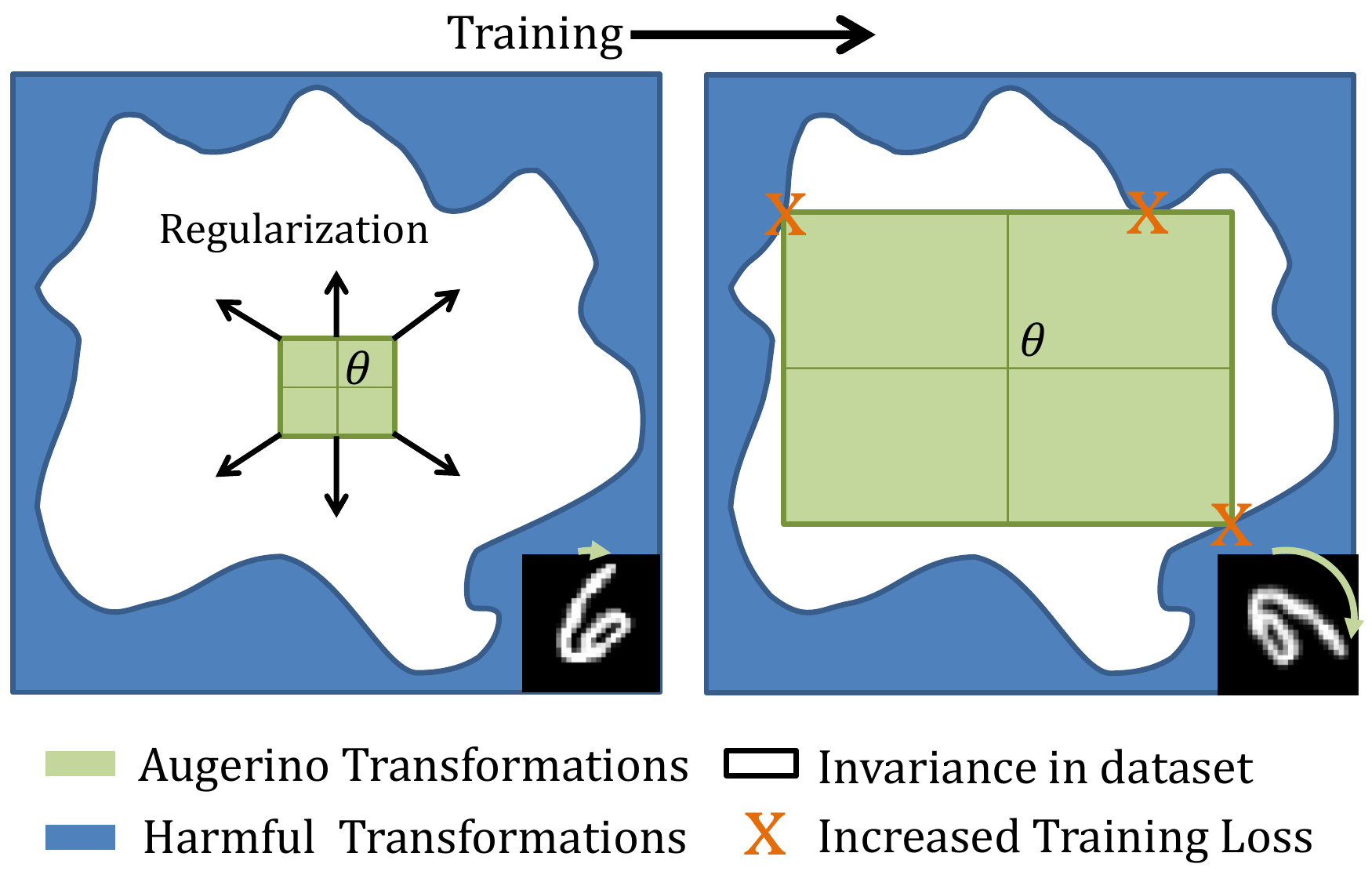} 
         \\
         (a) Augerino training
    \end{tabular}
    \hspace{-0.5cm}
    \begin{tabular}{c}
         \includegraphics[height=0.22\textwidth]{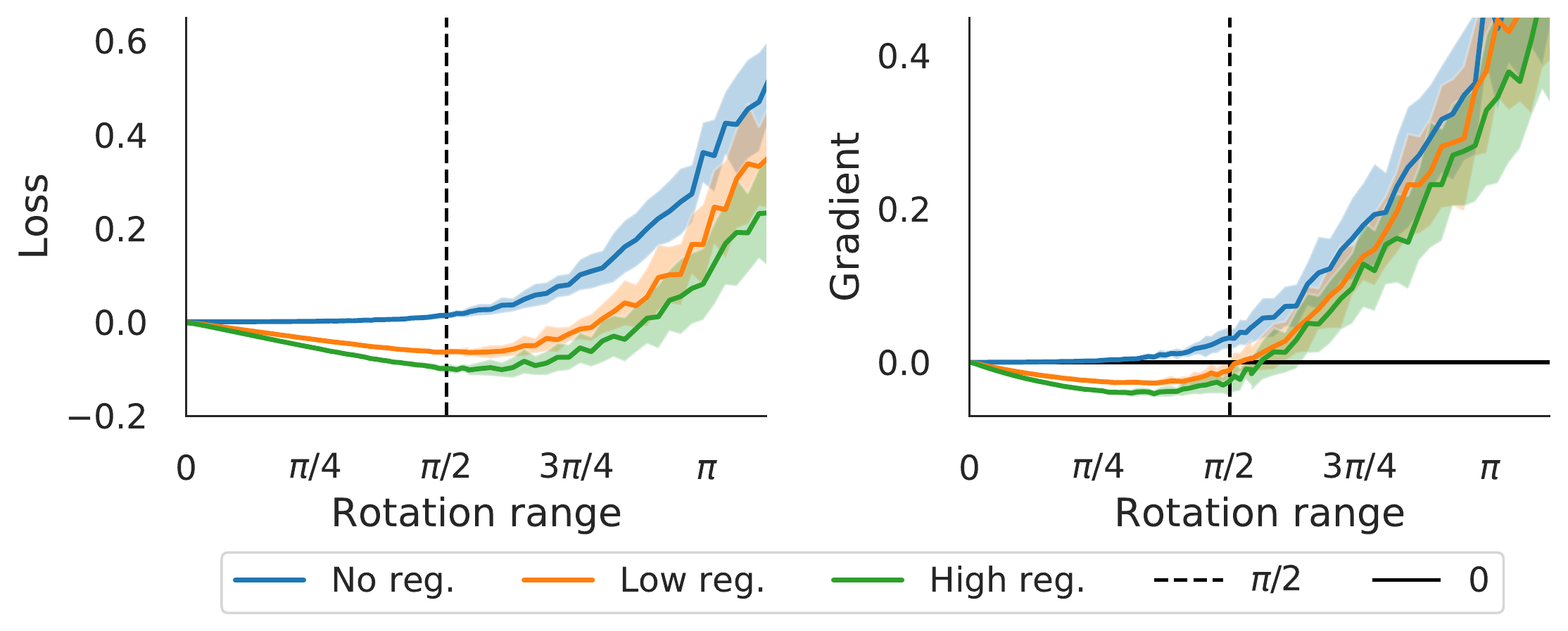} 
         \\
         (b) Loss function and Gradient
    \end{tabular}

    \caption{\textbf{(a):} A visualization of the space of possible transformations. Augerino expands to fill out the invariances in the dataset but is halted at the boundary where harmful transformations increase the training loss like rotating a 6 to a 9.
    \textbf{(b):} Loss value as a function of the rotation range applied to the input on the Mario and Iggy classification problem of Section \ref{sec:soft} and its derivative.
    Without regularization the loss is flat for augmentations within the range $[0, \pi/2]$ corresponding to the true rotational invariance range in the data, and grows sharply beyond this range.
    }
    \label{fig: understanding}
\end{figure*}

The conventional wisdom is that it is impossible to learn invariances directly from the training loss as invariances are constraints on the model which make it harder to fit the data. 
Given data that has invariance to some augmentation, the training loss will not be improved by widening our distribution over this augmentation, even if it helps generalization: we would want a model to be invariant to rotations of a `6' up until it looks more like a `9', but no invariance will achieve the same training loss. However, it is sufficient to add a simple regularization term to encourage the model to discover invariances. In practice we find that the final distribution over augmentations is insensitive to the level of regularization, and that even a small amount of regularization will enable Augerino to find wide distributions over augmentations that are consistent with the precise level of invariances in the data.

We illustrate the learning of invariances with Augerino in panel (a) of Figure \ref{fig: understanding}.
Suppose only a limited degree of invariance is present in the data, as in Section \ref{sec:soft}.
Then the training loss for the augmentation parameters will be flat for augmentations within the range of invariance present in the data (shown in white), and then will increase sharply beyond this range (corresponding region of Augerino parameters is shown in blue).
The regularized loss in Eq.~\eqref{eq:augerino_final_loss}
will push the model to increase the level of invariance within the flat region of the training loss, but will not push it beyond the degree of invariance present in the data unless the regularization strength is extreme.

\begin{figure}
    \centering
    \includegraphics[width=0.45\linewidth]{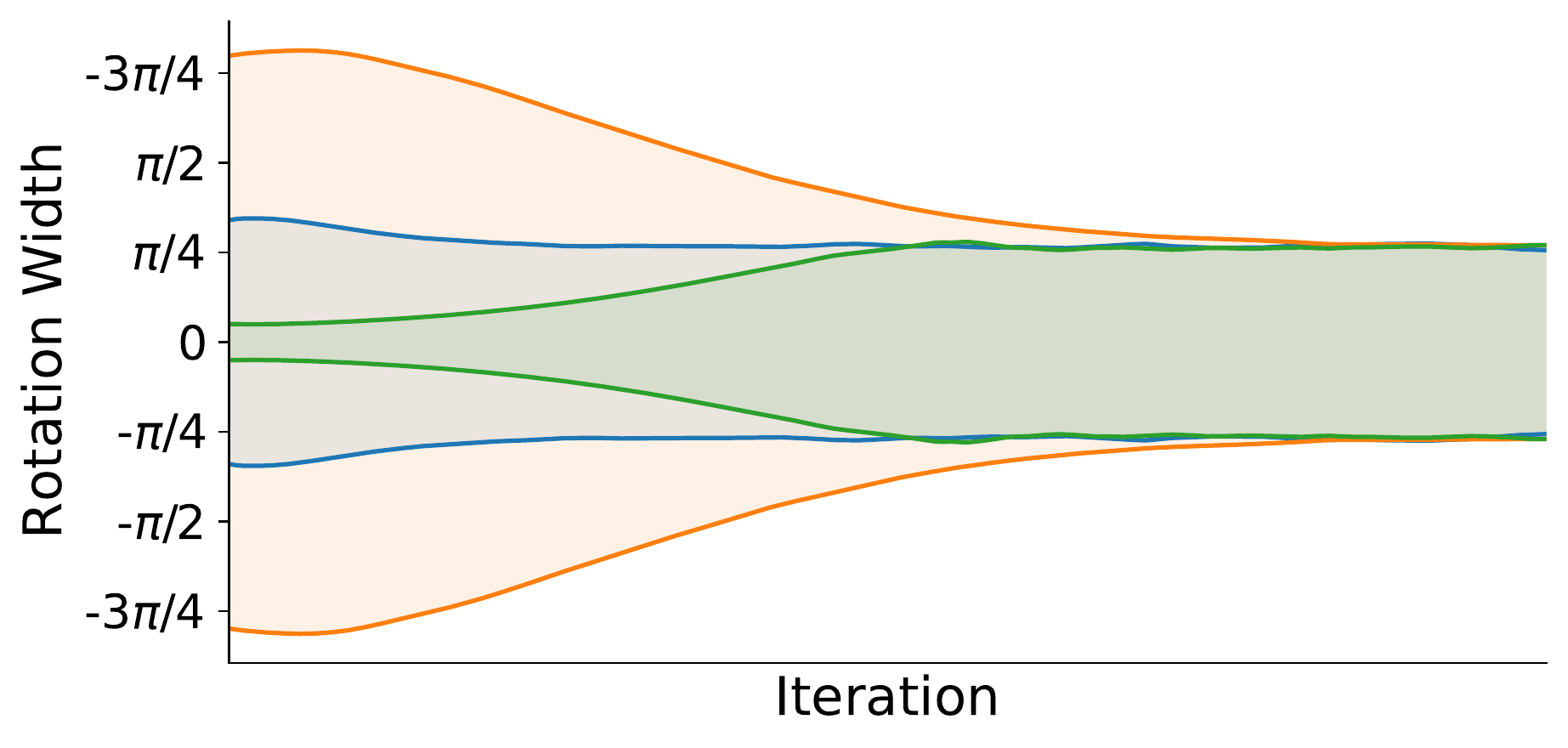}
    \caption{The distribution over rotation augmentations for the Mario and Iggy dataset over training iterations for various initializations. Regardless of whether we start with too wide, too narrow, or approximately the correct distribution over rotations, Augerino converges to the appropriate width.}
    \label{fig: mario-init}
\end{figure}

We demonstrate the effect described above for the Mario and Iggy classification problem of Section \ref{sec:soft} in panel (b) of Figure \ref{fig: understanding}.
We use a network trained with Augerino and visualize the loss and gradient with respect to the range of rotations applied to the input with and without regularization.
Without regularization, the loss is almost completely flat until the value of $\pi / 2$ which is the true degree of rotational invariance in the data.
With regularization we add an incentive for the model to learn larger values of the rotation range.  
Consequently, the loss achieves its optimum close to the optimal value of the parameter at $\pi / 2$ and then quickly grows beyond that value.
Figure \ref{fig: mario-init} displays the results of panel (b) of Figure \ref{fig: understanding} in action; gradient signals push augmentation distributions that are too wide down and too narrow up to the correct width.  

Incidentally, the Augerino solutions are substantially flatter than those obtained by standard training, as shown in Appendix \ref{app: cifar-flatness}, Figure \ref{fig: cifar10}, which may also make them more easily discoverable by procedures such as SGD. We also see that these solutions indeed provide better generalization. We provide further discussion of learning partial invariances with Augerino in Appendix~\ref{app: invariance}.

\section{Image Recognition}
\label{sec: boosting}
As Augerino learns a set of augmentations specific to a given dataset, we expect to see that Augerino is capable of boosting performance over applying any level of fixed augmentation. Using the CIFAR-$10$ dataset, we compare Augerino to training on data with $i)$ no augmentation, $ii)$ fixed, commonly applied augmentations, and $iii)$ the augmentations as given by Fast AutoAugment \citet{lim2019fast}. 

\begin{table}[h]
\small
    \centering
    \caption{Test accuracy for models trained on CIFAR-$10$ with different augmentations applied to the training data.}    
    \begin{tabular}{cccccc}
    \toprule
      & No Aug. & Fixed Aug. & Augerino ($4$ copies) & Augerino ($1$ copy) & Fast AA \\
     \midrule
     Test Accuracy & $90.60$ & $92.64$ & $\mathbf{93.81} \pm 0.002$ & $92.22 \pm 0.002$ & $92.65$ \\
     \bottomrule
    \end{tabular}
    \caption*{We compare models trained with no augmentation, a fixed commonly applied set of augmentations (including flipping, cropping, and color-standardization), Augerino, and Fast AutoAugment \citep{lim2019fast}. Augerino with $n_{copies}=4$ provides a boost in performance with minimal increased training time. 
    Error bars are reported as the standard deviation in accuracy for Augerino trained over $10$ trials.}
    \label{table:cifar}
\end{table}

Table \ref{table:cifar} shows that Augerino is competitive with advanced models that seek data-based augmentation schemes. The gains in performance are accompanied by notable simplifications in setup: we do not require a validation set and the augmentation is learned concurrently with training (there is no pre-processing to search for an augmentation policy). In Appendix \ref{app: cifar-flatness} we show that Augerino find \emph{flatter} solutions in the loss surface, which are known to generalize \citep{maddox2020rethinking}.
To further address the choice of regularization parameter, we train a number of models on CIFAR-$10$ with varying levels of regularization. 
In Figure \ref{fig: cifar10} we present the test accuracy of models for different regularization parameters along with the corresponding effective dimensionalities of the networks as a measure of the \emph{flatness} of the optimum found through training. 
\cite{maddox2020rethinking} shows that effective dimensionality can capture the flatness of optima in parameter space and is strongly correlated to generalization, with lower effective dimensionality implying flatter optima and better generalization.

The results of the experiment presented in Figure \ref{fig: cifar10} solidify Augerino's capability to boost performance on image recognition tasks as well as demonstrate that the inclusion of regularization is helpful, but not necessary to train accurate models. 
If the regularization parameter becomes too large, as can be seen in the rightmost violins of Figure \ref{fig: cifar10}, training can become unstable with more variance in the accuracy achieved.
We observe that while it is possible to achieve good results with no regularization, the inclusion of an inductive bias that we ought to include some invariances (by adding a regularization penalty) improves performance.

\section{Molecular Property Prediction}
\label{sec: molecules}
We test out our method on the molecular property prediction dataset QM9 \citep{blum, rupp} which consists of small inorganic molecules with features given by the coordinates of the atoms in 3D space and their charges. We focus on the HOMO task of predicting the energy of the highest occupied molecular orbital, and we learn Augerino augmentations in the space of affine transformations of the atomic coordinates in $\mathbb{R}^3$. We parametrize the transformation as before with a uniform distribution for each of the generators listed in Appendix \ref{app: generating-matrices}. We use the LieConv model introduced in \citet{finzi2020generalizing}, both with no equivariance (LieConv-Trivial) and 3D translational equivariance (LieConv-T$(3)$). We train the models for 500 epochs on MAE (additional training details are given in \ref{app: training-details}) and report the test performance in Table \ref{table:qm9}. Augerino performs much better than using no augmentations and is competitive with the hand chosen random rotation and translation augmentation ($\mathrm{SE}(3)$) that incorporates domain knowledge about the problem. 
We detail the learned distribution over affine transformations in Appendix \ref{sec:qm9_aug}. Augerino is useful both for the non equivariant LieConv-Trivial model as well as the translationally equivariant LieConv-T(3) model, suggesting that Augerino can complement architectural equivariance.
\begin{table}[h]
\vspace{-1mm}
    \centering
    \small
    \caption{Test MAE (in meV) on QM9 tasks trained with specified augmentation.}
    \begin{tabular}{cccccccc}
    \multicolumn{1}{c}{}&
    \multicolumn{3}{c}{HOMO (meV)}&
    \multicolumn{3}{c}{LUMO (meV)}\\
    \toprule
      &No Aug.& Augerino &SE(3)&No Aug.& Augerino &SE(3) \\
     \midrule
     LieConv-Trivial & $52.7$ & $38.3$ & $\mathbf{36.5}$ &$43.5$ & $33.7$&$\mathbf{29.8}$ \\
     LieConv-T(3)& $34.2$ & $33.2$ & $\mathbf{30.2}$ &$30.1$ & $26.9$&$\mathbf{25.1}$ \\
     \bottomrule
    \end{tabular}
    \label{table:qm9}
    \vspace{-5mm}
\end{table}

\section{Semantic Segmentation}
\label{sec:semseg}

In Section \ref{sec:equivariance} we showed how Augerino can be extended to equivariant problems.
In Semantic Segmentation the targets are perfectly aligned with the inputs and the network should be equivariant to any transformations present in the data.
To test Augerino in equivariant learning setting we construct rotCamVid, a variation of the CamVid dataset \citep{BrostowSFC:ECCV08, BrostowFC:PRL2008} where all the training and test points are rotated by a random angle (see Appendix Figure \ref{fig: semseg}). 
For any fixed image we always use the same rotation angle, so no two copies of the same image with different rotations are present in the data.
We use the FC-Densenet segmentation architecture \citep{jegou2017one}.
We train Augerino with a Gaussian distribution over random rotations and translations.

In Appendix Figure \ref{fig: semseg} we visualize the training data and learned augmentations for Augerino.
Augerino is able to successfully recover rotational augmentation while matching the performance of the baseline. 
For further details, please see Appendix \ref{sec: app_semseg}.

\section{Color-Space Augmentations}
\label{sec:color}

In the previous sections we have focused on learning spatial invariances with Augerino.
Augerino is general and can be applied to arbitrary differentiable input transformations.
In this section, we demonstrate that Augerino can learn color-space invariances.

We consider two color-space augmentations: brightness adjustments and contrast adjustments. 
Each of these can be implemented as simple differentiable transformations to the RGB values of the input image (for details, see Appendix \ref{sec: app_colorspace}).
We use Augerino to learn a uniform distribution over the brightness and contrast adjustments on STL-10 \citep{pmlr-v15-coates11a} using the $13$-layer CNN architecture (see Section \ref{sec: shades}).
For both Augerino and the baseline model, we use standard spatial data augmentation: random translations, flips and cutout \citep{devries2017improved}.
The baseline model achieves $89.0 \pm 0.35 \%$ accuracy where the mean and standard deviation are computed over $3$ independent runs.
The Augerino model achieves a slightly higher $89.7 \pm 0.3\%$ accuracy and learns to be invariant to noticeable brightness and contrast changes in the input image (see Appendix Figure \ref{fig: stl-colors}). 

\section{Conclusion}

We have introduced \emph{Augerino}, a framework that can be seamlessly deployed with standard model architectures to learn symmetries from training data alone, and improve generalization. 
Experimentally, we see that Augerino is capable of recovering ground truth invariances, including \emph{soft} invariances, ultimately discovering an interpretable representation of the dataset. Augerino's ability to recover interpretable and accurate distributions over augmentations leads to increased performance over both task-specific specialized baselines and competing data-based augmentation schemes on a variety of tasks including molecular property prediction, image segmentation, and classification.

\newpage
\section*{Broader Impacts}

Our work is largely methodological and we anticipate that Augerino will primarily see use within the machine learning community. Augerino's ability to uncover invariances present within the data, \emph{without} modifying the training procedure and with a very plug-and-play design that is compatible with any network architecture makes it an appealing method to be deployed widely. We hope that learning invariances from data is an avenue that will see continued inquiry and that Augerino will motivate further exploration.

\section*{Acknowledgements}

This research is supported by an Amazon Research Award, Facebook Research, Amazon Machine Learning Research Award, NSF I-DISRE 193471, NIH R01 DA048764-01A1, NSF IIS-1910266, and NSF 1922658 NRT-HDR: FUTURE Foundations, Translation, and Responsibility for Data Science.

\nocite{*}
\bibliographystyle{apalike}
\bibliography{refs}

\begin{thebibliography}{}

\bibitem[Athiwaratkun et~al., 2019]{athiwaratkun2018improving}
Athiwaratkun, B., Finzi, M., Izmailov, P., and Wilson, A.~G. (2019).
\newblock There are many consistent explanations of unlabeled data: Why you
  should average.
\newblock {\em ICLR}.

\bibitem[Bekkers, 2020]{bekkers2019b}
Bekkers, E.~J. (2020).
\newblock B-spline cnns on lie groups.
\newblock In {\em International Conference on Learning Representations}.

\bibitem[Blum and Reymond, 2009]{blum}
Blum, L.~C. and Reymond, J.-L. (2009).
\newblock 970 million druglike small molecules for virtual screening in the
  chemical universe database {GDB-13}.
\newblock {\em J. Am. Chem. Soc.}, 131:8732.

\bibitem[Brostow et~al., 2008a]{BrostowFC:PRL2008}
Brostow, G.~J., Fauqueur, J., and Cipolla, R. (2008a).
\newblock Semantic object classes in video: A high-definition ground truth
  database.
\newblock {\em Pattern Recognition Letters}.

\bibitem[Brostow et~al., 2008b]{BrostowSFC:ECCV08}
Brostow, G.~J., Shotton, J., Fauqueur, J., and Cipolla, R. (2008b).
\newblock Segmentation and recognition using structure from motion point
  clouds.
\newblock In {\em ECCV (1)}, pages 44--57.

\bibitem[Coates et~al., 2011]{pmlr-v15-coates11a}
Coates, A., Ng, A., and Lee, H. (2011).
\newblock An analysis of single-layer networks in unsupervised feature
  learning.
\newblock In Gordon, G., Dunson, D., and Dudík, M., editors, {\em Proceedings
  of the Fourteenth International Conference on Artificial Intelligence and
  Statistics}, volume~15 of {\em Proceedings of Machine Learning Research},
  pages 215--223, Fort Lauderdale, FL, USA. PMLR.

\bibitem[Cohen and Welling, 2016a]{cohen2016group}
Cohen, T. and Welling, M. (2016a).
\newblock Group equivariant convolutional networks.
\newblock In {\em International conference on machine learning}, pages
  2990--2999.

\bibitem[Cohen et~al., 2019]{cohen2019general}
Cohen, T.~S., Geiger, M., and Weiler, M. (2019).
\newblock A general theory of equivariant cnns on homogeneous spaces.
\newblock In {\em Advances in Neural Information Processing Systems}, pages
  9142--9153.

\bibitem[Cohen and Welling, 2016b]{cohen2016steerable}
Cohen, T.~S. and Welling, M. (2016b).
\newblock Steerable cnns.
\newblock {\em arXiv preprint arXiv:1612.08498}.

\bibitem[Cubuk et~al., 2019]{cubuk2019autoaugment}
Cubuk, E.~D., Zoph, B., Mane, D., Vasudevan, V., and Le, Q.~V. (2019).
\newblock Autoaugment: Learning augmentation strategies from data.
\newblock In {\em Proceedings of the IEEE conference on computer vision and
  pattern recognition}, pages 113--123.

\bibitem[Dao et~al., 2019]{dao2019kernel}
Dao, T., Gu, A., Ratner, A.~J., Smith, V., De~Sa, C., and R{\'e}, C. (2019).
\newblock A kernel theory of modern data augmentation.
\newblock {\em Proceedings of machine learning research}, 97:1528.

\bibitem[DeVries and Taylor, 2017]{devries2017improved}
DeVries, T. and Taylor, G.~W. (2017).
\newblock Improved regularization of convolutional neural networks with cutout.

\bibitem[Falorsi et~al., 2019]{falorsi2019reparameterizing}
Falorsi, L., de~Haan, P., Davidson, T.~R., and Forr{\'e}, P. (2019).
\newblock Reparameterizing distributions on lie groups.
\newblock {\em arXiv preprint arXiv:1903.02958}.

\bibitem[Finzi et~al., 2020]{finzi2020generalizing}
Finzi, M., Stanton, S., Izmailov, P., and Wilson, A.~G. (2020).
\newblock Generalizing convolutional neural networks for equivariance to lie
  groups on arbitrary continuous data.
\newblock {\em arXiv preprint arXiv:2002.12880}.

\bibitem[Hataya et~al., 2019]{hataya2019faster}
Hataya, R., Zdenek, J., Yoshizoe, K., and Nakayama, H. (2019).
\newblock Faster autoaugment: Learning augmentation strategies using
  backpropagation.
\newblock {\em arXiv preprint arXiv:1911.06987}.

\bibitem[Hinton and Salakhutdinov, 2008]{hinton2008using}
Hinton, G.~E. and Salakhutdinov, R.~R. (2008).
\newblock Using deep belief nets to learn covariance kernels for gaussian
  processes.
\newblock In {\em Advances in neural information processing systems}, pages
  1249--1256.

\bibitem[Huang et~al., 2017]{huang2017deep}
Huang, Z., Wan, C., Probst, T., and Van~Gool, L. (2017).
\newblock Deep learning on lie groups for skeleton-based action recognition.
\newblock In {\em Proceedings of the IEEE conference on computer vision and
  pattern recognition}, pages 6099--6108.

\bibitem[Jaderberg et~al., 2015]{jaderberg2015spatial}
Jaderberg, M., Simonyan, K., Zisserman, A., et~al. (2015).
\newblock Spatial transformer networks.
\newblock In {\em Advances in neural information processing systems}, pages
  2017--2025.

\bibitem[J{\'e}gou et~al., 2017]{jegou2017one}
J{\'e}gou, S., Drozdzal, M., Vazquez, D., Romero, A., and Bengio, Y. (2017).
\newblock The one hundred layers tiramisu: Fully convolutional densenets for
  semantic segmentation.
\newblock In {\em Proceedings of the IEEE conference on computer vision and
  pattern recognition workshops}, pages 11--19.

\bibitem[Kingma and Welling, 2013]{kingma2013auto}
Kingma, D.~P. and Welling, M. (2013).
\newblock Auto-encoding variational bayes.
\newblock {\em arXiv preprint arXiv:1312.6114}.

\bibitem[Laine and Aila, 2016]{laine2016temporal}
Laine, S. and Aila, T. (2016).
\newblock Temporal ensembling for semi-supervised learning.
\newblock {\em arXiv preprint arXiv:1610.02242}.

\bibitem[Laptev et~al., 2016]{laptev2016ti}
Laptev, D., Savinov, N., Buhmann, J.~M., and Pollefeys, M. (2016).
\newblock Ti-pooling: transformation-invariant pooling for feature learning in
  convolutional neural networks.
\newblock In {\em Proceedings of the IEEE conference on computer vision and
  pattern recognition}, pages 289--297.

\bibitem[Larochelle et~al., 2007]{rotmnist}
Larochelle, H., Erhan, D., Courville, A., Bergstra, J., and Bengio, Y. (2007).
\newblock An empirical evaluation of deep architectures on problems with many
  factors of variation.
\newblock In {\em Proceedings of the 24th International Conference on Machine
  Learning}, ICML ’07, page 473–480, New York, NY, USA. Association for
  Computing Machinery.

\bibitem[LeCun et~al., 1998]{lecun1998convolutional}
LeCun, Y., Bengio, Y., et~al. (1998).
\newblock Convolutional networks for images, speech, and time series, the
  handbook of brain theory and neural networks.

\bibitem[Li et~al., 2020]{li2020dada}
Li, Y., Hu, G., Wang, Y., Hospedales, T., Robertson, N.~M., and Yang, Y.
  (2020).
\newblock Dada: Differentiable automatic data augmentation.
\newblock {\em arXiv preprint arXiv:2003.03780}.

\bibitem[Liang et~al., 2019]{liang2019darts}
Liang, H., Zhang, S., Sun, J., He, X., Huang, W., Zhuang, K., and Li, Z.
  (2019).
\newblock Darts+: Improved differentiable architecture search with early
  stopping.
\newblock {\em arXiv preprint arXiv:1909.06035}.

\bibitem[Lim et~al., 2019]{lim2019fast}
Lim, S., Kim, I., Kim, T., Kim, C., and Kim, S. (2019).
\newblock Fast autoaugment.
\newblock In {\em Advances in Neural Information Processing Systems}, pages
  6662--6672.

\bibitem[Liu et~al., 2018]{liu2018darts}
Liu, H., Simonyan, K., and Yang, Y. (2018).
\newblock Darts: Differentiable architecture search.
\newblock {\em arXiv preprint arXiv:1806.09055}.

\bibitem[MacKay, 2003]{mackay2003information}
MacKay, D.~J. (2003).
\newblock {\em Information theory, inference and learning algorithms}.
\newblock Cambridge university press.

\bibitem[Maddox et~al., 2020]{maddox2020rethinking}
Maddox, W.~J., Benton, G., and Wilson, A.~G. (2020).
\newblock Rethinking parameter counting in deep models: Effective
  dimensionality revisited.
\newblock {\em arXiv preprint arXiv:2003.02139}.

\bibitem[Marcos et~al., 2017]{marcos2017rotation}
Marcos, D., Volpi, M., Komodakis, N., and Tuia, D. (2017).
\newblock Rotation equivariant vector field networks.
\newblock In {\em Proceedings of the IEEE International Conference on Computer
  Vision}, pages 5048--5057.

\bibitem[Minka, 2001]{minka2001automatic}
Minka, T.~P. (2001).
\newblock Automatic choice of dimensionality for pca.
\newblock In {\em Advances in neural information processing systems}, pages
  598--604.

\bibitem[Nintendo, 1990]{smw}
Nintendo (1990).
\newblock Super mario world.

\bibitem[Raj et~al., 2017]{raj2017local}
Raj, A., Kumar, A., Mroueh, Y., Fletcher, T., and Sch{\"o}lkopf, B. (2017).
\newblock Local group invariant representations via orbit embeddings.
\newblock In {\em Artificial Intelligence and Statistics}, pages 1225--1235.

\bibitem[Rupp et~al., 2012]{rupp}
Rupp, M., Tkatchenko, A., M\"uller, K.-R., and von Lilienfeld, O.~A. (2012).
\newblock Fast and accurate modeling of molecular atomization energies with
  machine learning.
\newblock {\em Physical Review Letters}, 108:058301.

\bibitem[Sosnovik et~al., 2019]{sosnovik2019scale}
Sosnovik, I., Szmaja, M., and Smeulders, A. (2019).
\newblock Scale-equivariant steerable networks.
\newblock {\em arXiv preprint arXiv:1910.11093}.

\bibitem[Tran et~al., 2017]{tran2017bayesian}
Tran, T., Pham, T., Carneiro, G., Palmer, L., and Reid, I. (2017).
\newblock A bayesian data augmentation approach for learning deep models.
\newblock In {\em Advances in neural information processing systems}, pages
  2797--2806.

\bibitem[van~der Wilk et~al., 2018]{van2018learning}
van~der Wilk, M., Bauer, M., John, S., and Hensman, J. (2018).
\newblock Learning invariances using the marginal likelihood.
\newblock In {\em Advances in Neural Information Processing Systems}, pages
  9938--9948.

\bibitem[Weiler and Cesa, 2019]{weiler2019general}
Weiler, M. and Cesa, G. (2019).
\newblock General e (2)-equivariant steerable cnns.
\newblock In {\em Advances in Neural Information Processing Systems}, pages
  14334--14345.

\bibitem[Worrall and Welling, 2019]{worrall2019deep}
Worrall, D. and Welling, M. (2019).
\newblock Deep scale-spaces: Equivariance over scale.
\newblock In {\em Advances in Neural Information Processing Systems}, pages
  7364--7376.

\bibitem[Worrall et~al., 2017]{worrall2017harmonic}
Worrall, D.~E., Garbin, S.~J., Turmukhambetov, D., and Brostow, G.~J. (2017).
\newblock Harmonic networks: Deep translation and rotation equivariance.
\newblock In {\em Proceedings of the IEEE Conference on Computer Vision and
  Pattern Recognition}, pages 5028--5037.

\bibitem[Xu et~al., 2019]{xu2019pc}
Xu, Y., Xie, L., Zhang, X., Chen, X., Qi, G.-J., Tian, Q., and Xiong, H.
  (2019).
\newblock Pc-darts: Partial channel connections for memory-efficient
  differentiable architecture search.
\newblock {\em arXiv preprint arXiv:1907.05737}.

\bibitem[Zhang et~al., 2019]{zhang2019dada}
Zhang, X., Wang, Z., Liu, D., and Ling, Q. (2019).
\newblock Dada: Deep adversarial data augmentation for extremely low data
  regime classification.
\newblock In {\em ICASSP 2019-2019 IEEE International Conference on Acoustics,
  Speech and Signal Processing (ICASSP)}, pages 2807--2811. IEEE.

\bibitem[Zhou et~al., 2017]{zhou2017oriented}
Zhou, Y., Ye, Q., Qiu, Q., and Jiao, J. (2017).
\newblock Oriented response networks.
\newblock In {\em Proceedings of the IEEE Conference on Computer Vision and
  Pattern Recognition}, pages 519--528.

\end{thebibliography}

\newpage\null
\appendix
\addcontentsline{toc}{section}{Appendix} 
\part{Appendix} 
\parttoc

Here we present additional details and experimental results. 
In Section \ref{app: invariance} we give a further discussion of the formation of our invariant model. 
Section \ref{app: generating-matrices} gives the form of the generating matrices for the Lie group and the corresponding transformations to which they give rise.
In Section \ref{sec: app_semseg} we provide details regarding the experimental setup and results in applying Augerino to image segmentation. In Section \ref{app: training-details} we give the full training details for the experiments of Sections \ref{sec: shades} and \ref{sec: boosting}. In Section \ref{sec: app_colorspace} we expand on the details of the color-space augmentation experiment given in Section \ref{sec:color} in the main text. 
Section \ref{sec:qm9_aug} expands on the molecular property prediction experiments of Section \ref{sec: molecules}, showing the learned augmentations and giving further details regarding the experimental setup.
Finally Section \ref{app: cifar-flatness} explains how Augerino aids in finding solutions that generalize well through looking at the \emph{effective dimensionality} of the training solutions \citep{maddox2020rethinking}.

\section{Forming The Invariant Model}
\label{app: invariance}

We form a model that is approximately invariant to transformations in $\textrm{supp}(\mu_{\theta}) = \mathcal{S}$ by taking the expectation over transformations $g\sim \mu_{\theta}$:
\begin{equation}\label{eqn: aug-expec-app}
    \bar{f}(x) = \E_{g\sim \mu_\theta}f(gx).
\end{equation}
If $\mu_{\theta}$ is uniform over the full span of a transformation, such as rotations in $[-\pi, \pi]$, 
then $\bar{f}(x)$ will be exactly invariant with respect to that transformation.
In cases where $\mathcal{S}$ has only partial support over transformations.
Equation \eqref{eqn: aug-expec-app} alone does not imply invariance. For example, let $\mu_{\theta}$ be a uniform distribution over rotations in $[-\pi/2, \pi/2]$. Then for an input image $x$ and and an input $x' = r_{\pi/2}x$, i.e. the image $x$ rotated by $\pi/2$ radians, we have
\begin{align*}
    \bar{f}(x) &= \int_{-\pi/2}^{\pi/2}f(r_{\phi}x) d\phi \\
    \bar{f}(x') &= \int_{-\pi/2}^{\pi/2}f(r_{\phi}x') d\phi = \int_{0}^{\pi}f(r_{\phi}x) d\phi, 
\end{align*}
therefore without additional properties on $f$, we \emph{cannot} guarantee that $\bar{f}(x) = \bar{f}(x')$. This behaviour is in contrast to the case of having a complete invariance where the support of $\mu_\theta$ is closed over transformations.

However, even in these cases of partial support over invariances, the training procedure still leads to invariant or nearly invariant models (also referred to as \emph{insensitivity} in \citet{van2018learning}). This empirical fact can be naturally understood from the perspective of data augmentation.
Once we iterate through the training set many times, then for each input $x$ the network $\bar{f}$ will have been trained on inputs $gx$ for many $g \sim \mu_\theta$.  If our network achieves near $0$ training loss, as is typical for image problems, then we will have a network which predicts the same correct label for each input $gx$ with $g \sim \mu_{\theta}$, giving a network $\bar{f}$ that is approximately invariant to the correct augmentations. In practice, the network will generalize this insensitivity to transformations on unseen test data.

In particular, Augerino learns the maximal possible augmentations that do not hurt training performance.
For example, suppose we observe rotations of the digit `6' in the range $[-\pi/4, \pi/4]$ from the vertical. Augerino will learn rotation invariance up to $\pi/4$, as rotating further will move some of the observations below the upper half plane, where they may be more correctly labelled as `9'.
Once $\mu_{\theta}$ has converged to $[-\pi/4, \pi/4]$, $\bar{f}$ will be trained to correctly classify observations of the digit `6' rotated over the upper half plane, giving approximate invariance to any rotation in $[-\pi/4, \pi/4]$.

\section{Lie Group Generators}
\label{app: generating-matrices}

The six Lie group generating matrices for affine transformations in 2D are,
\begin{align}
    \label{eqn: generators}
    \begin{split}
      G_1 &= 
      \begin{bmatrix}
           0&0&1  \\
           0&0&0  \\
           0&0&0  \\
      \end{bmatrix},
      \quad
      G_2 = 
      \begin{bmatrix}
           0&0&0  \\
           0&0&1  \\
           0&0&0  \\
      \end{bmatrix},\quad
      G_3 = 
      \begin{bmatrix}
           0&-1&0  \\
           1&0&0  \\
           0&0&0  \\
      \end{bmatrix},\\
      G_4 &= 
      \begin{bmatrix}
           1&0&0  \\
           0&1&0  \\
           0&0&0  \\
      \end{bmatrix}, \quad
      G_5 = 
      \begin{bmatrix}
           1&0&0  \\
           0&-1&0  \\
           0&0&0  \\
      \end{bmatrix}, \quad
      G_6 = 
      \begin{bmatrix}
           0&1&0  \\
           1&0&0  \\
           0&0&0  \\
      \end{bmatrix}.
    \end{split}
\end{align}
Applying the exponential map to these matrices produces affine matrices that can be used to transform images. In order, these matrices correspond to translations in $x$, translations in $y$, rotations, scaling in $x$, scaling in $y$, and shearing. 

\section{Semantic Segmentation: Details}
\label{sec: app_semseg}

\begin{figure*}%
    \centering
    \begin{subfigure}{0.22\textwidth}
        \includegraphics[width=\linewidth]{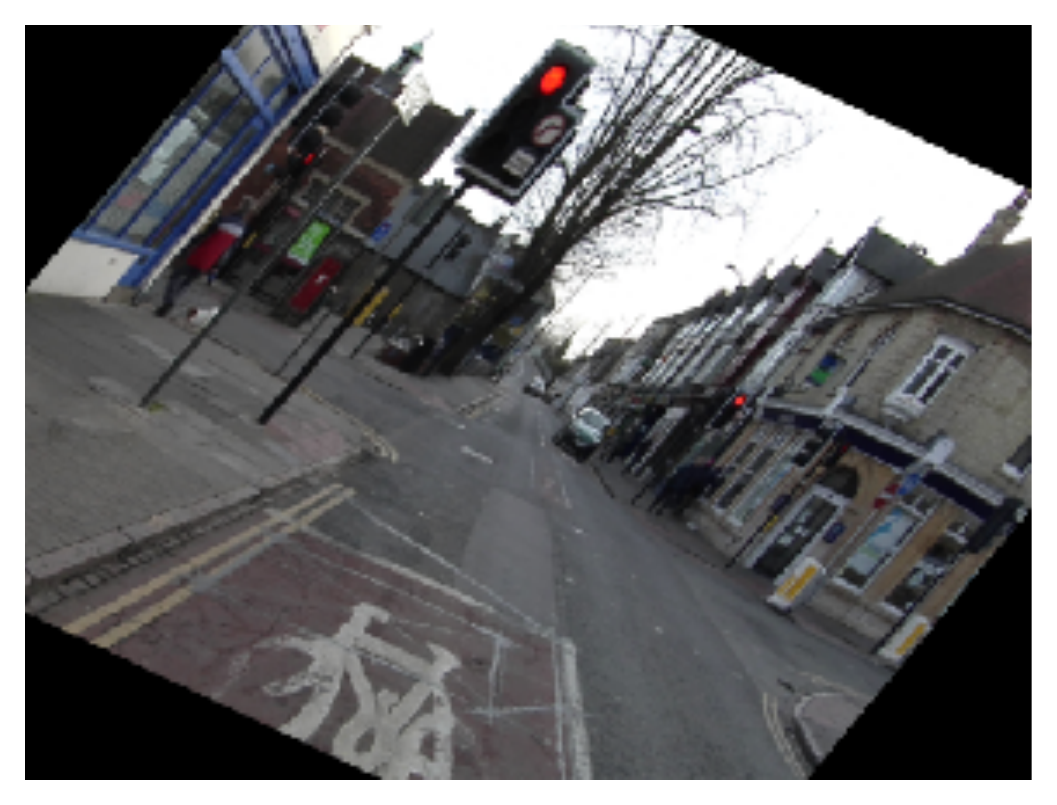}
        \caption{Original Data}
    \end{subfigure}
    ~
    \begin{subfigure}{0.22\textwidth}
        \includegraphics[width=\linewidth]{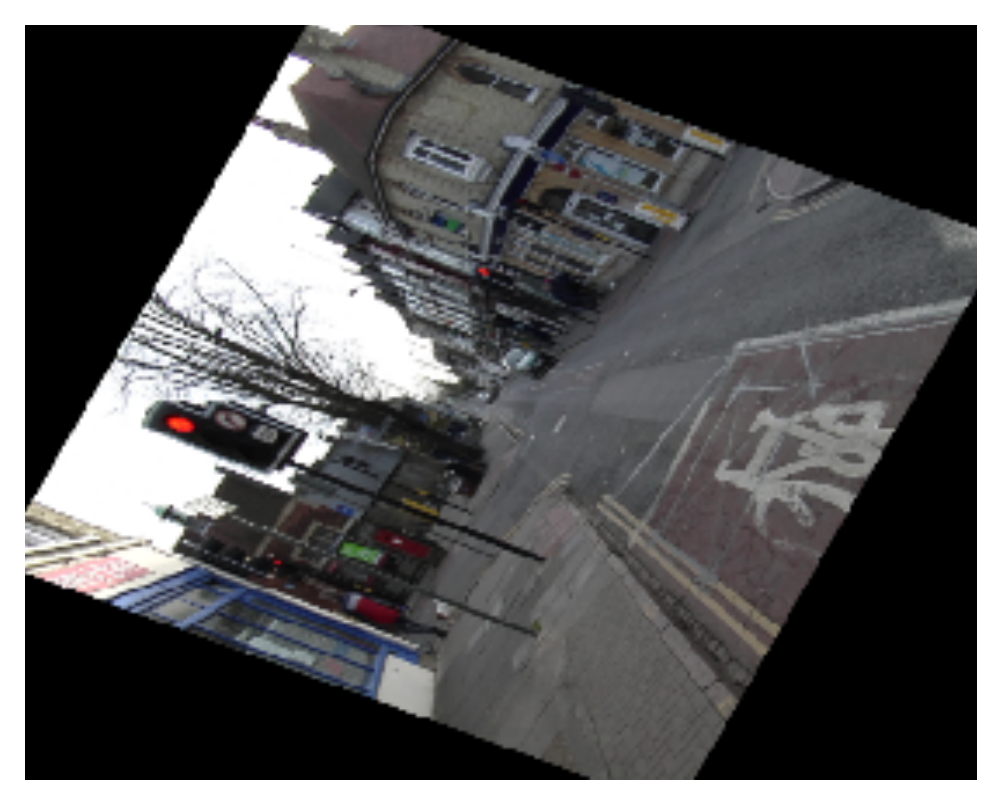}       \caption{Augerino Sample}
    \end{subfigure}
    ~
    \begin{subfigure}{0.22\textwidth}
        \includegraphics[width=\linewidth]{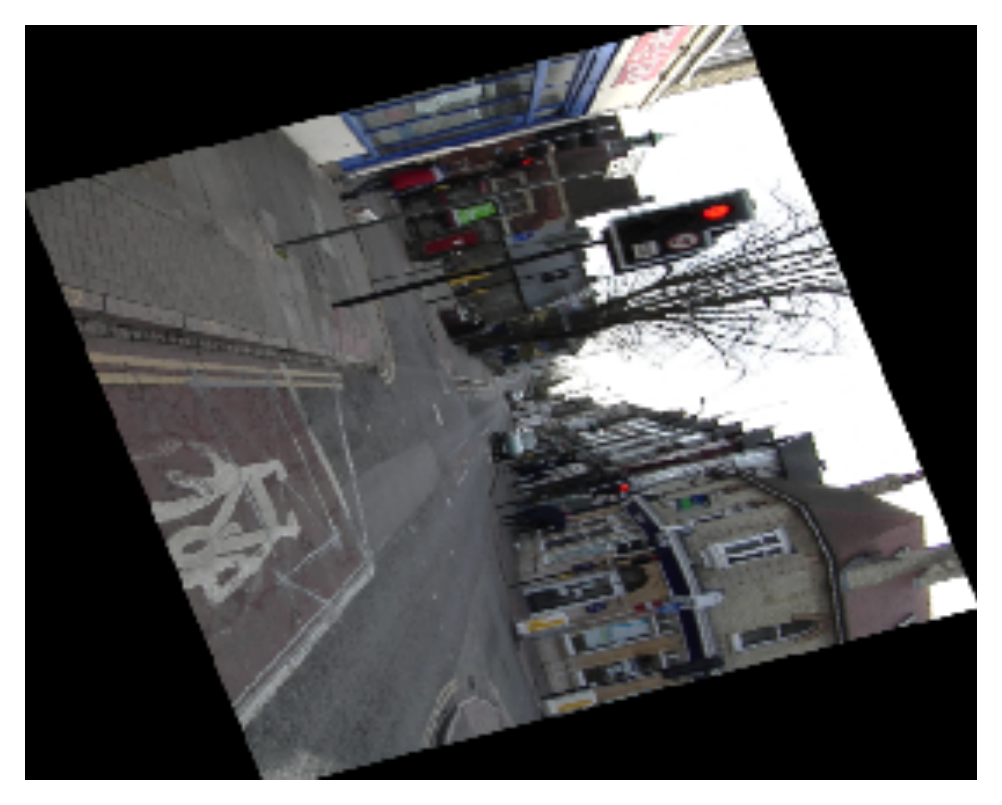}
        \caption{Augerino Sample}
    \end{subfigure}
    ~
    \begin{subfigure}{0.22\textwidth}
        \includegraphics[width=\linewidth]{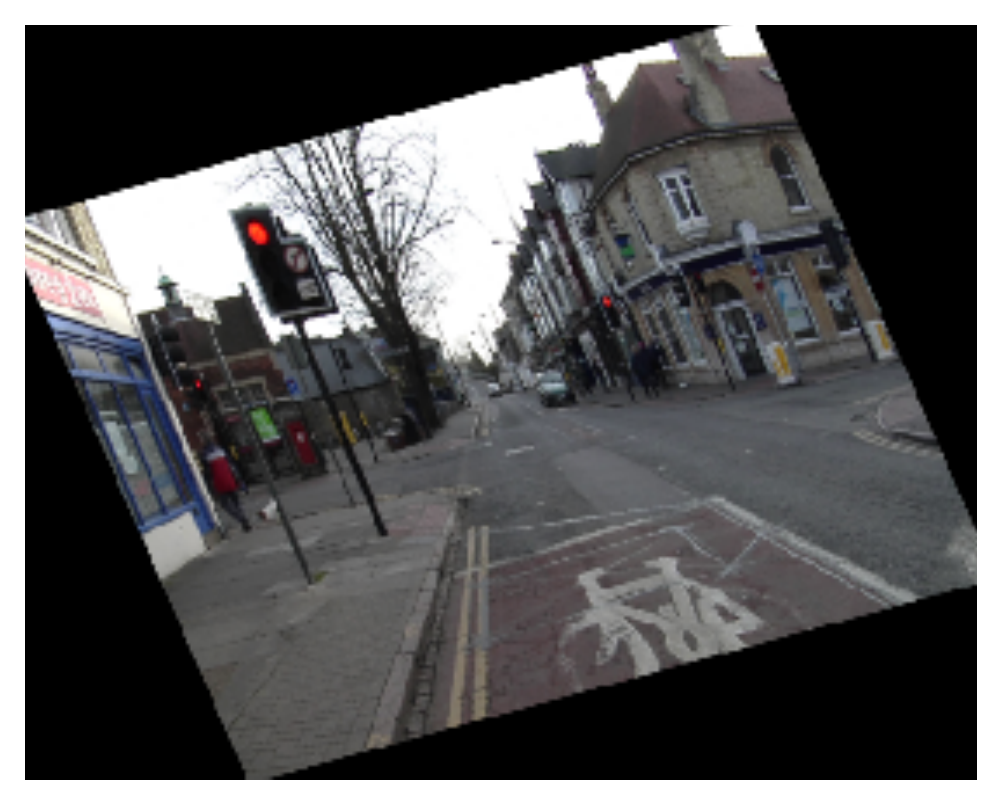}
        \caption{Augerino Sample}
    \end{subfigure}
    \caption{Augmentations learned by Augerino on the rotCamVid dataset.
    \textbf{(a)}: original data from rotCamVid; \textbf{(b)-(d)}: three random samples of augmentations from the learned augerino distribution.
    Augerino learns to be invariant to rotations but not translations. 
    }
    \label{fig: semseg}
\end{figure*}

In Section \ref{sec:semseg}, we apply Augerino to semantic segmentation on the rotCamVid dataset (see Figure \ref{fig: semseg}).

To generate the rotCamVid dataset, we rotate all images in the CamVid by a random angle, analogously to the rotMNIST dataset \citep{rotmnist}.
We note that rotCamVid only contains a single rotated copy of each image, which is not the same as applying rotational augmentation during training. 
When computing the training loss and test acccuracy, we ignore the padding pixels which appear due to rotating the image.

For the segmentation experiment we used the simpler augmentation distribution covering rotations and translations instead of the affine transformations (Section \ref{sec:affine}).
We use a Gaussian parameterization of the distribution:
\begin{equation}
    t = (t_1, t_2, t_3) \sim \mathcal N(\mu, \Sigma),
    \quad
    A(t) = 
    \begin{bmatrix}
        \cos(t_1) & -\sin(t_1) & 2 \cdot t_2  / (w + h) \\
        \sin(t_1) & \cos(t_1) & 2 \cdot t_3  / (w + h)
    \end{bmatrix},
\end{equation}
where $\mu, \Sigma$ are trainable parameters, and $A(t)$ is the affine transformation matrix for the random sample $t$; $w$ and $h$ are the width and height of the image.

Augerino achieves pixel-wise segmentation accuracy of $69.8\%$ while the baseline model with standard augmentation achieves $68.7\%$.

\section{Training Details}
\label{app: training-details}

\paragraph{Network Training Hyperparameters}
We train the networks in Sections \ref{sec: shades} and \ref{sec: boosting} for $200$ epochs, using an initial learning rate of $0.01$ with a cosine learning rate schedule and a batch size of $128$. We use the cross entropy loss function for all classification tasks, and mean squared error for all regression tasks except for QM9 where we use mean absolute error.

\paragraph{Train- and Test-Time Augmentations}

In Algorithm \ref{alg:augerino} we include a term \emph{ncopies} that denotes the number of sampled augmentations during training. We find that we can achieve strong performance with Augerino, with minimally increased training time, by setting \emph{ncopies} to $1$ at train-time and then applying multiple augmentations by increasing \emph{ncopies} at \emph{test-time}. Thus we train using a single augmentation for each input, and then apply multiple augmentations at test-time to increase accuracy, as seen in Table \ref{table:cifar}.

\section{Color-Space Augmentations: Details}
\label{sec: app_colorspace}

\begin{figure*}%
    \centering
    \begin{subfigure}{0.22\textwidth}
        \includegraphics[width=\linewidth]{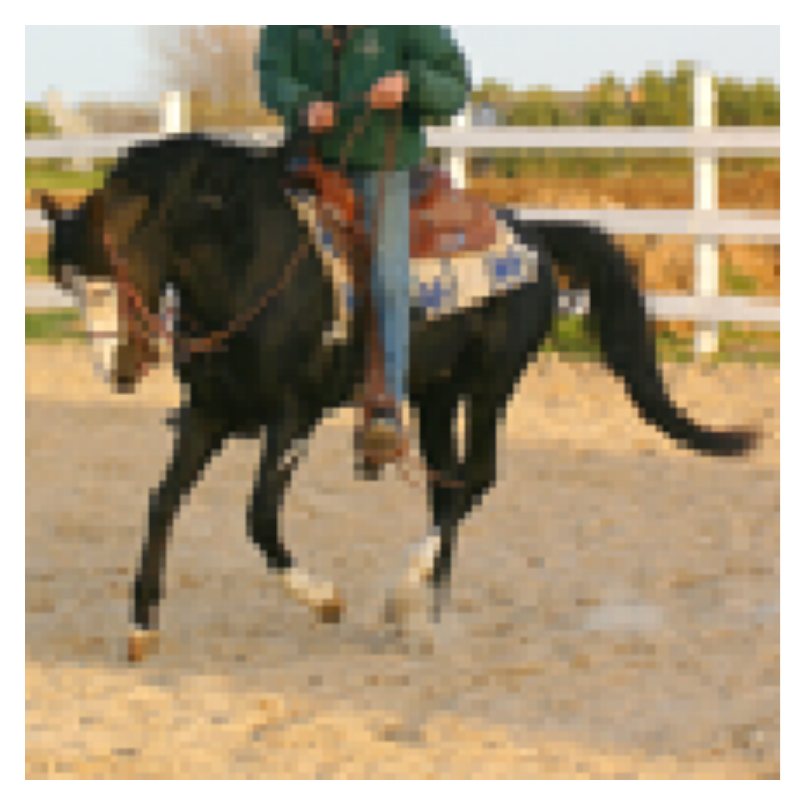}
        \caption{Original Data}
    \end{subfigure}
    ~
    \begin{subfigure}{0.22\textwidth}
        \includegraphics[width=\linewidth]{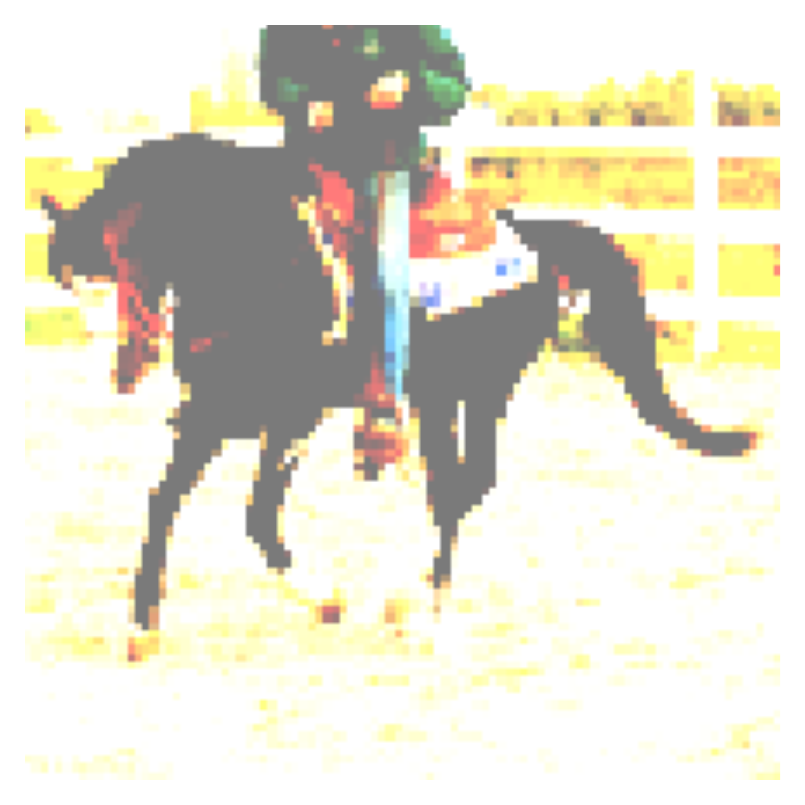}
        \caption{Augerino Sample}
    \end{subfigure}
    ~
    \begin{subfigure}{0.22\textwidth}
        \includegraphics[width=\linewidth]{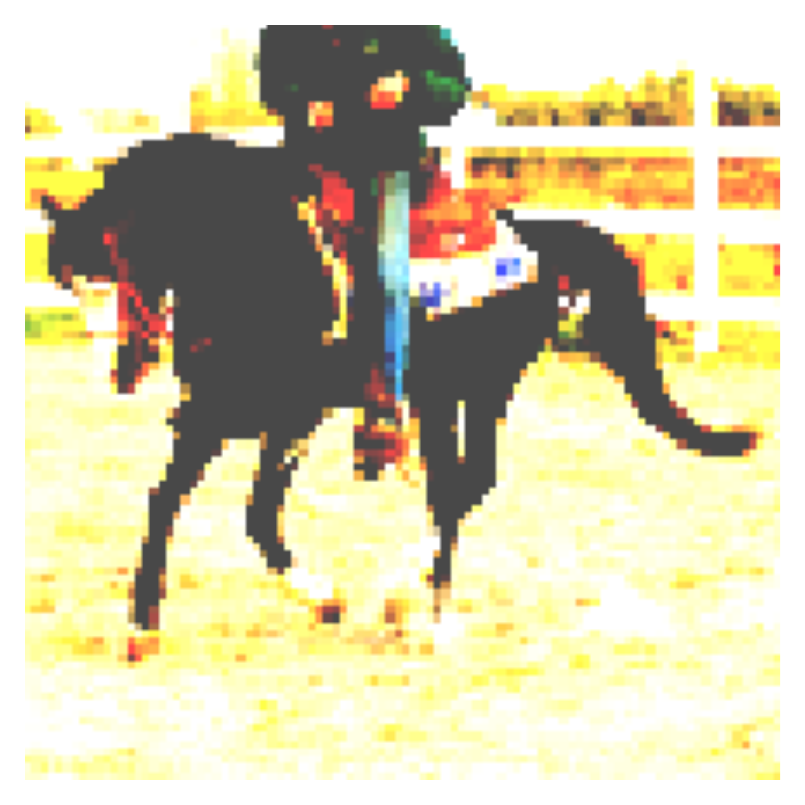}
        \caption{Augerino Sample}
    \end{subfigure}
    ~
    \begin{subfigure}{0.22\textwidth}
        \includegraphics[width=\linewidth]{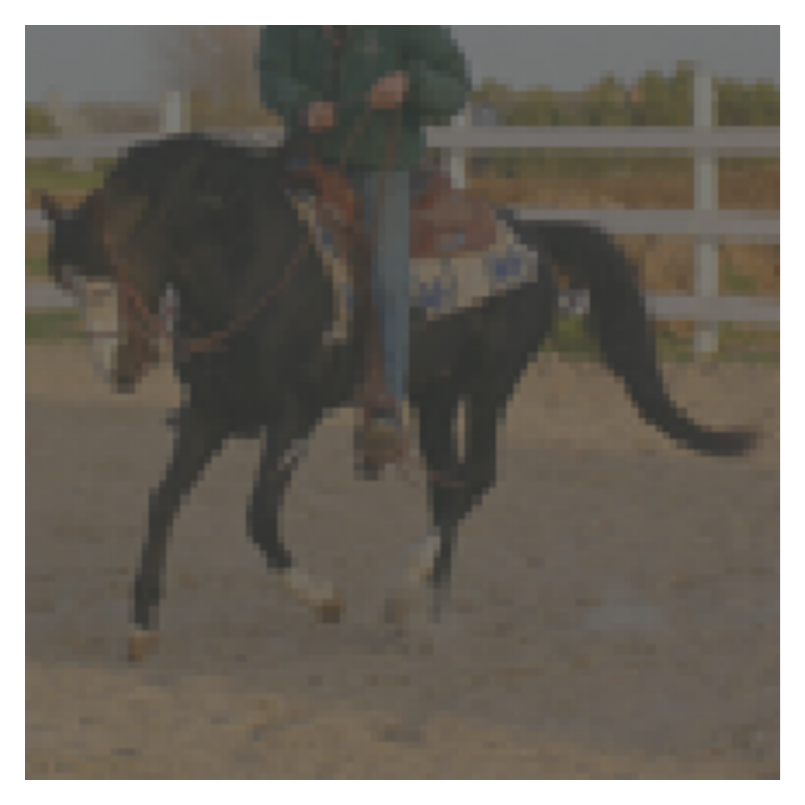}
        \caption{Augerino Sample}
    \end{subfigure}
    \caption{Color-space augmentation distribution learned by Augerino.
    \textbf{(a)}: original data from STL-10; \textbf{(b)-(d)}: three random samples of augmentations from the learned augerino distribution.
    Augerino learns to be invariant to a broad range of color and contrast adjustments while matching the performance of the baseline. 
    }
    \label{fig: stl-colors}
\end{figure*}

In Section \ref{sec:color}, we apply Augerino to learning color-space invariances on the STL-10 dataset. 
We consider two transformations: 
\begin{itemize}
    \item Brightness adjustment by a value $t$ transforms the intensity $c$ in each channel additively:
    \begin{equation}
        \label{eq:brightness}
        c' = \max(\min(c + t, 255), 0). 
    \end{equation}
    Positive $t$ increases, and negative $t$ decreases brightness.
    
    \item Contrast adjustment by a value $t$ transforms the intensity $c$ in each channel as follows\footnote{\url{https://www.dfstudios.co.uk/articles/programming/image-programming-algorithms/image-processing-algorithms-part-5-contrast-adjustment/}}:
    \begin{equation}
        \label{eq:contrast}
        c' = \max\bigg(\min\bigg( \frac{259 \cdot (t +  255)}{255 \cdot(259 - t)} \cdot (c - 128) + 128, ~~255 \bigg), 0\bigg )
    \end{equation}
    
\end{itemize}

We apply brightness and contrast adjustments sequentially and independently from each other.
We learn the range of a uniform distribution over the values $t$ in \eqref{eq:brightness}, \eqref{eq:contrast}.
The learned data augmentation strategy is visualized in Figure \ref{fig: stl-colors}.

\section{QM9 Experiment}\label{sec:qm9_aug}
We reproduce the training details from \citet{finzi2020generalizing}. Affine transformations in 3d, there are 9 generators, 3 for translation, 3 for rotation, 2 for squeezing and 1 for scaling, a straightforward extension of those listed in equation \ref{eqn: generators} to 3 dimensions. Like before, we parametrize the bounds on the uniform distribution for each of these generators. We use a regularization strength of $10^{-3}$.

\section{Width of Augerino Solutions}
\label{app: cifar-flatness}
\begin{figure}
    \centering
    \includegraphics[width=\linewidth]{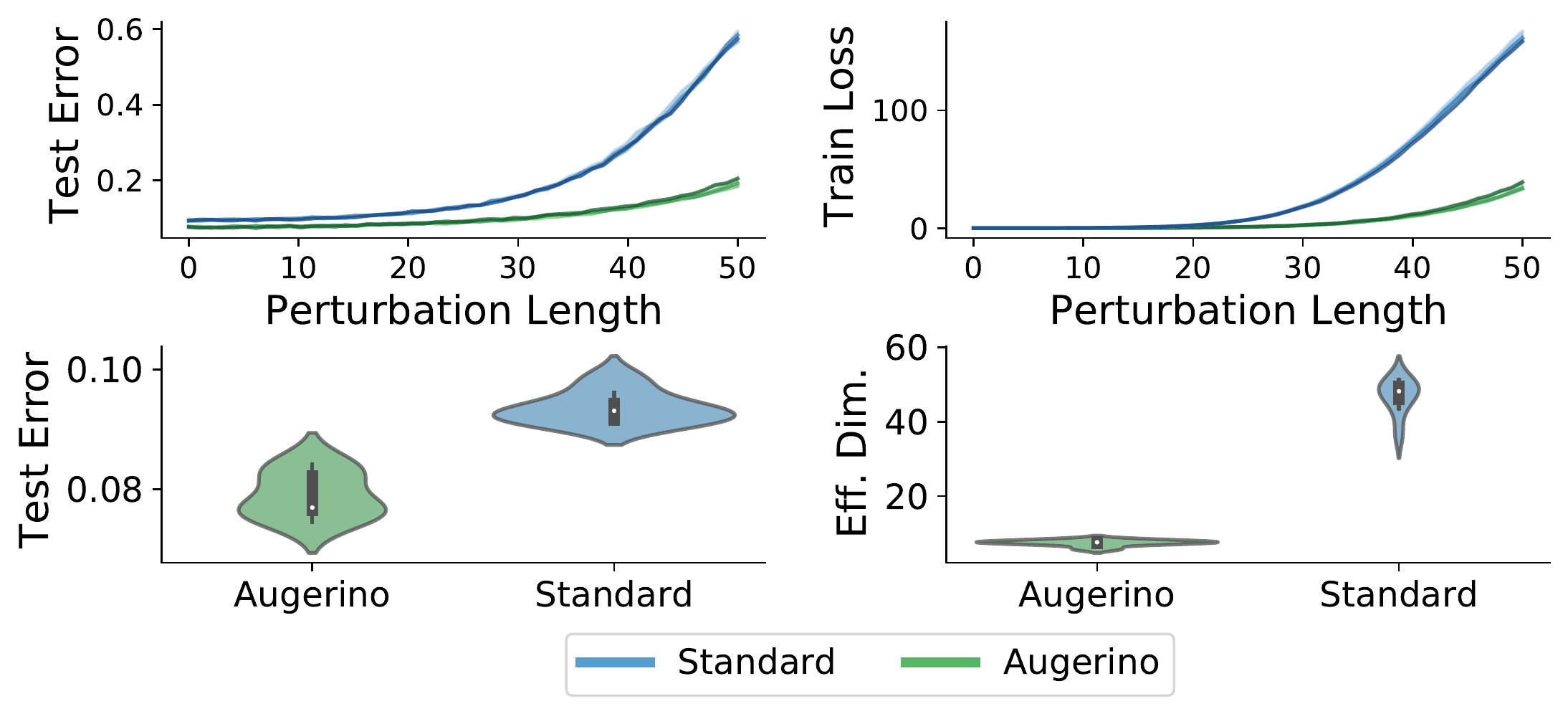}
    \caption{\textbf{Top:} Test error and train loss as a function of perturbation lengths along random rays from the SGD found training solution for models. Each curve represents a different ray. 
    \textbf{Bottom:} Test error and effective dimensionality for models trained on CIFAR-$10$. Results from $8$ random initializations are presented violin-plot style where width represents the kernel density estimate at the corresponding $y$-value.}
    \label{fig: cifar10}
\end{figure}

To help explain the increased generalization seen in using Augerino, we train $10$ models on CIFAR-$10$ both with and without Augerino. 
In Figure \ref{fig: cifar10} we present the test error of both types of models for along with the corresponding effective dimensionalities and sensitivity to parameter perturbations of the networks as a measure of the \emph{flatness} of the optimum found through training. 
\citet{maddox2020rethinking} shows that effective dimensionality can capture the flatness of optima in parameter space and is strongly correlated to generalization, with lower effective dimensionality implying flatter optima and better generalization.
Overall we see that Augerino enables networks to find much flatter solutions in the loss surface, corresponding to better compressions of the data and better generalization.

\end{document}